\documentclass[a4paper]{article}

\usepackage{INTERSPEECH2020}
\usepackage{amsmath,mathtools}
\usepackage{url}

\title{\textit{Mere account mein kitna balance hai?} - On building voice enabled Banking Services for Multilingual Communities}
\name{Akshat Gupta, Sai Krishna Rallabandi and Alan W Black}
\address{Department of Electrical and Computer Engineering, Carnegie Mellon University\\
  Language Technologies Institute, Carnegie Mellon University}
\email{ akshatgu@andrew.cmu.edu, \{srallaba,awb\} @cs.cmu.edu}

\begin{document}

\maketitle
\begin{abstract}
Tremendous progress in speech and language processing has brought language technologies closer to daily human life. Voice technology has the potential to act as a horizontal enabling layer across all aspects of digitization. It is especially beneficial to rural communities in scenarios like a pandemic. In this work\footnote{This is an extended abstract.} we present our initial exploratory work towards one such direction - building voice enabled banking services for multilingual societies. Speech interaction for typical banking transactions in multilingual communities involves the presence of filled pauses and is characterized by Code Mixing. Code Mixing is a phenomenon where lexical items from one language are embedded in the utterance of another. Therefore speech systems deployed for banking applications should be able to process such content. In our work we investigate various training strategies for building speech based intent recognition systems. We present our results using a Naive Bayes classifier on approximate acoustic phone units using the Allosaurus library \cite{li2020universal}. 
\end{abstract}

\section{Data Collection Tool}
\label{data_collection}

We have created a dataset of multilingual queries using a dummy banking app \cite{bankingapp}. It involves a two stage setup. In the first stage, speech data is crowdsourced from fluent Hindi speakers. Each speaker plays the role of a `user' and interacts with the automated banking system.  This resulted in 100+ task-based dialogs in Hinglish, with five distinct intents. In the second stage, each speaker is asked to acoustically translate their interaction into another language. This way, we obtain pseudo parallel data in two languages from the same speaker. We believe that creating a pseudo parallel dataset will allow us to design semi supervised approaches in the future. Our dataset currently has speech data from six Indian languages - Hindi, Marathi, Gujarati, Punjabi, Telugu and Bhojpuri.

\section{Methodology and Results}
We chose spoken intent classification as the first task. For this task, we have 25 utterances containing speech samples of 11 people, out of which 4 were female. We have 5 intents in the dataset - \textbf{Send Money} (11 utterances), \textbf{Check Balance} (9 utterances), \textbf{Check Last Transaction} (3 utterances), \textbf{Withdraw Money} and \textbf{Deposit Money} (1 utterance each) .

We first convert audio into phones using the Allosaurus library\cite{li2020universal}. These phones are then used for intent classification. We employ Naive Bayes' Classifier with add-1 smoothing and absolute discounting. We use cross validation where we leave out 2 audio samples for testing. The intents \textit{Withdraw Money} and \textit{Deposit Money} were not used for testing as we only have one sample for both, but are included in the training set. Thus the testing is only done for three intents, but an intent could be classified into any of the 5 classes. The results for Naive Bayes Classifier for add-1 smoothing are shown in Table \ref{resultsadd1}.

\begin{table}
\begin{center}
 \begin{tabular}{||c c c||} 
 \hline
    \textbf{Model} &  \textbf{\# Unique N-grams} & \textbf{ Test Accuracy } \\ [0.5ex] 
 \hline\hline
 Unigram & 38 & 0.56   \\ 
 \hline
 Bigram & 292 & 0.48 \\
 \hline
 Trigram & 543 & 0.17 \\
 \hline
\end{tabular}
\caption{N-gram classification accuracy with Add-1 smoothing}
\label{resultsadd1}
\end{center}
\end{table}

\begin{table}
\begin{center}
 \begin{tabular}{||c c c||} 
 \hline
    \textbf{Model} & \textbf{Delta} & \textbf{ Test Accuracy } \\ [0.5ex] 
 \hline\hline
 Unigram  & 5 & 0.69   \\ 
 \hline
 Bigram & 1 & 0.61 \\
 \hline
 Trigram & 1 & 0.30 \\
 \hline
 Combination & (5, 1, 1) & \textbf{0.83} \\
 \hline
\end{tabular}
\caption{Classification accuracy with Absolute Discounting}
\label{resultsAD}
\end{center}
\end{table}

It can be observed that the accuracy decreases with increasing N for the different N-gram models. We hypothesize that this happens due to the relatively small size of the dataset in our initial exploratory work. We posit the distribution characteristics to be more uniform across N-grams in our future experiments with full dataset.

Further we have also performed absolute discounting with different delta values for each of the Ngram models. The test accuracies improved significantly. We also combined unigram, bigram and trigram models with equal weights with their respective best performing delta values, which gave us the best result as shown in Table \ref{resultsAD}.

\section{Conclusion}
In this paper, we present our initial exploration towards acoustic intent classification using a Naive Bayes classifier on approximate acoustic phone units. Our results indicate that our system can be employed to build real life banking applications in multilingual scenarios.

\bibliographystyle{IEEEtran}

\bibliography{mybib}

\end{document}